\documentclass[10pt,twocolumn,letterpaper]{article}

\usepackage{cvpr}              

\usepackage{graphicx}
\usepackage{amsmath}
\usepackage{amssymb}
\usepackage{booktabs}

\usepackage[misc]{ifsym}

\usepackage{mathrsfs}
\usepackage{enumitem}
\usepackage{algpseudocode}
\usepackage{multirow}
\usepackage{mathrsfs}
\usepackage{booktabs}
\usepackage[linesnumbered,algoruled,boxed,lined]{algorithm2e}

\usepackage[pagebackref=false,breaklinks=true,letterpaper=true,colorlinks,urlcolor=blue,citecolor=blue,linkcolor=blue,bookmarks=false]{hyperref}

\usepackage[capitalize]{cleveref}
\crefname{section}{Sec.}{Secs.}
\Crefname{section}{Section}{Sections}
\Crefname{table}{Table}{Tables}
\crefname{table}{Tab.}{Tabs.}

\def\cvprPaperID{7382} 
\def\confName{CVPR}
\def\confYear{2022}

\begin{document}

\title{Exploring Frequency Adversarial Attacks for Face Forgery Detection}

\author{
Shuai Jia$^1$\quad 
Chao Ma$^1$\thanks{~Corresponding author.}\quad
Taiping Yao$^2$\quad 
Bangjie Yin$^2$\quad
Shouhong Ding$^2$\quad 
Xiaokang Yang$^1$ \\
$^1$ MoE Key Lab of Artificial Intelligence, AI Institute, Shanghai Jiao Tong University  \\
$^2$ Youtu Lab, Tencent\\
{\tt\small \{jiashuai,chaoma,xkyang\}@sjtu.edu.cn} \\
{\tt\small \{taipingyao,ericshding\}@tencent.com, \tt\small jamesyin10@gmail.com}
}

\maketitle

\begin{abstract}
Various facial manipulation techniques have drawn serious public concerns in morality, security, and privacy. Although existing face forgery classifiers achieve promising performance on detecting fake images, these methods are vulnerable to adversarial examples with injected imperceptible perturbations on the pixels.  Meanwhile, many face forgery detectors always utilize the frequency diversity between real and fake faces as a crucial clue. In this paper, instead of injecting adversarial perturbations into the spatial domain, we propose a frequency adversarial attack method against face forgery detectors. Concretely, we apply discrete cosine transform (DCT) on the input images and introduce a fusion module to capture the salient region of adversary in the frequency domain. Compared with existing adversarial attacks (e.g. FGSM, PGD) in the spatial domain, our method is more imperceptible to human observers and does not degrade the visual quality of the original images. Moreover, inspired by the idea of meta-learning, we also propose a hybrid adversarial attack that performs attacks in both the spatial and frequency domains. Extensive experiments indicate that the proposed method fools not only the spatial-based detectors but also the state-of-the-art frequency-based detectors effectively. In addition, the proposed frequency attack enhances the transferability across face forgery detectors as black-box attacks. 
\end{abstract}
\vspace{-0.15in}

\section{Introduction}

\newcommand{\figdir}{figures}
\def\swtwo{0.43\linewidth}
\begin{figure}[t]
    \small
	\begin{center}
			\begin{tabular}{cc}
				\includegraphics[width=\swtwo]{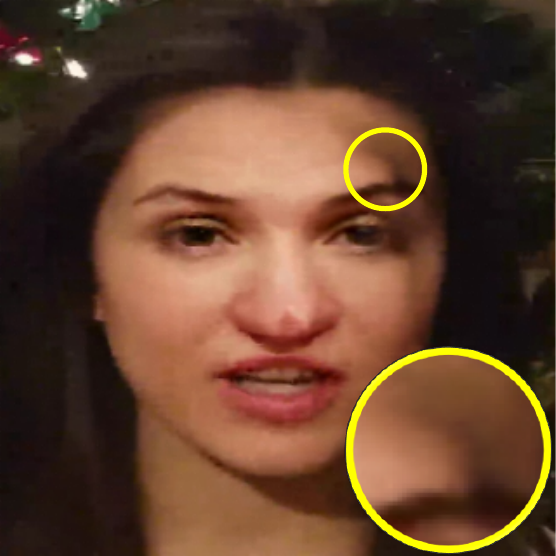}&
				\includegraphics[width=\swtwo]{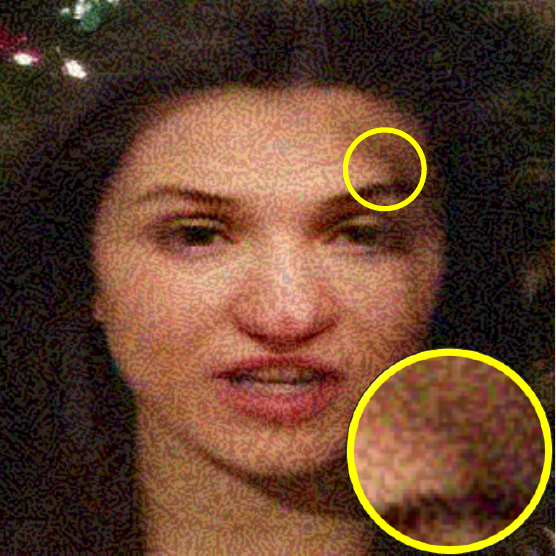}\\
				\vspace{0.05in}
				(a) Original image  & (b)  FGSM~\cite{xie-cvpr19-difgsm}\\
				\includegraphics[width=\swtwo]{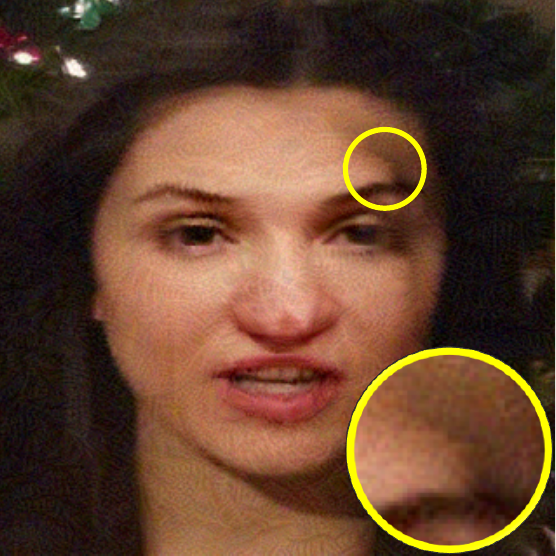}&
				\includegraphics[width=\swtwo]{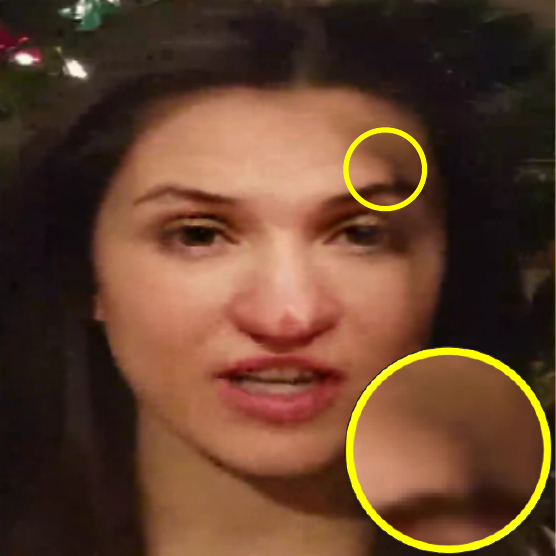}\\
				(c) PGD~\cite{madry-iclr18-pgd} & (d) Ours\\
			\end{tabular}
		\vspace{-0.20in}
	\end{center}
\caption{Illustration of adversarial examples generated by FGSM~\cite{goodfellow-2014-fgsm}, PGD~\cite{madry-iclr18-pgd} and our method. The original image is classified as a fake face by the face forgery detector. After implementing these attacks, the adversarial examples are misclassified as real faces. Compared with FGSM~\cite{goodfellow-2014-fgsm} and PGD~\cite{madry-iclr18-pgd}, our method associated with the frequency adversarial attack generates more natural perturbations, where the image quality of the adversarial example is much closer to the original image.}
\label{fig:intro}
\vspace{-0.05in}
\end{figure}

With the rapid development of generative adversarial network (GAN), face forgery generation attracts increasing attention, such as Deepfake~\cite{deepfake}, FaceSwap~\cite{facesawp}, Face2Face~\cite{thies-cvpr16-face2face}, and NeuralTextures~\cite{thies-tog19-neural}. These techniques derive plenty of interesting applications, for instance, trying on makeup virtually and editing faces in the film industry. However, despite the positive aspect, face forgery generation may be maliciously abused, causing serious problems of security and privacy. Therefore, it is essential to design face forgery detection methods to distinguish the manipulated face from the real one.

Various face forgery detectors~\cite{dang-cvpr20-detection,afchar-wifs18-mesonet,rossler--iccv19-faceforensics++,cozzolino-acmmm17-recasting} are proposed to learn the decision boundary between real and fake faces and achieve significant performance on multiple datasets~\cite{rossler-2018-ff++,dolhansky2019-dfdc}. However, existing methods are vulnerable to the adversarial examples, which leaves a serious backdoor for the security of detectors. For instance, a forged face image that is classified correctly as fake by adding adversarial perturbations can fool the detector to make a wrong decision as real. Existing works~\cite{li-cvpr21-exploring,neekhara-cvprw21-adversarial,carlini-cvprw20-evading,gandhi-ijcnn20-adversarial,hussain-wacv21-adversarial} have explored the robustness of face forgery detection methods, but these methods add adversarial perturbations or patches on the original images, which are easily recognized by human eyes. In brief, the adversarial examples aim to fool a face forgery detector, while the objective of face forgery generation is to fool humans. An implicit attack that fools humans and detectors at the same time brings out a more serious problem of security. Meanwhile, more and more works~\cite{qian-eccv20-thinking,chen-aaai21-local} consider the frequency diversity between real and fake faces as the essential clues for face forgery detection. It inspires us to conduct the adversarial attack in the frequency domain to boost the transferablility across various detectors.

To address the above issues, we propose a frequency adversarial attack method to add adversarial perturbations in the frequency domain. First, we apply discrete cosine transform (DCT) to transfer the input images into the frequency domain. Specifically, we utilize a fusion module to slightly modify the energy in different frequency bands via the adversarial loss. The indirect injection of adversary into frequency domain avoids the redundant noise of attacks in the spatial domain (e.g., FGSM~\cite{xie-cvpr19-difgsm}, PGD~\cite{madry-iclr18-pgd}) 
and does not degrade the visual quality of original images. After that, we apply inverse DCT back to the spatial domain and obtain the final adversarial examples.  For face forgery detectors, some existing methods~\cite{sun-cvpr21-spa2,zhao-cvpr21-spa1} only consider the noise pattern in the spatial domain to detect the fake faces, while others~\cite{qian-eccv20-thinking,liu-cvpr21-fre1,luo-cvpr21-fre2} utilize the frequency information as a clue. Moreover, some methods~\cite{chen-aaai21-local,masi-eccv20-two1,li-cvpr21-two2} combine the discriminative features from both domains to learn the boundary between real and fake faces. Therefore, in order to enhance the generalization of the proposed attack method, we propose a hybrid adversarial attack to integrate the spatial adversarial attack and frequency adversarial attack into a whole framework. Inspired by the idea of meta-learning~\cite{nichol2018first}, we alternately optimize the perturbations based on the adversarial gradients in different domains. The compatible ensemble of adversarial attacks can reserve the virtues of attacks in both domains. Adversarial examples with different attacks are illustrated in Figure~\ref{fig:intro}.

Our main contributions can be summarized as follows.
\begin{itemize} [noitemsep,nolistsep]
	\item For the task of face forgery detection, we propose a novel adversarial attack method to generate perturbations in the frequency domain. Compared with the previous attacks, our method generates more imperceptible perturbations for human observers. 
	\item To further boost the transferability of the attack, we propose a hybrid adversarial attack based on the strategy of meta-learning to simultaneously perform attacks on the spatial and frequency domain.
	\item We perform the proposed method both on the spatial-based face forgery detectors and the state-of-the-art frequency-based detectors. Extensive experiments on benchmarks demonstrate the effectiveness of our attack under both white-box and black-box settings.
\end{itemize}

\section{Related Work}
In this section, we briefly introduce the development of face forgery generation and detection. Besides, we review recent adversarial attack methods, especially for face forgery detection.

\subsection{Face Forgery Generation}
Face forgery generation~\cite{karras-cvpr19-stylegan,garrido-cvpr14-automatic,karras-iclr18-pggan} aims to craft the face image that is authentic in the eyes of human beings, which brings numerous productive applications, e.g., virtual shopping, online education, film production, etc. In summary, face forgery generation can be divided into four categories~\cite{mirsky2020creation}: reenactment, replacement, editing and synthesis. One typical application of reenactment is to use one's expression or mouth to drive another one, e.g., Recycle-GAN~\cite{bansal-eccv18-recycle}, STGAN~\cite{liu-cvpr2019-stgan}. FaceSwap~\cite{facesawp} is the most common type of replacement.  Averbuch-Elor et al.~\cite{averbuch-tog17-bringing} animate the expressiveness of the subject through 2D warps and transfer it to the target automatically. Face2Face~\cite{thies-cvpr16-face2face} considers the facial expressions as under-constrained problems to transfer the deformation between source and target. Editing and synthesis are used to add or remove ones' attributes consisting of hair, glasses, age, makeup, etc. In this paper, we choose the fake face images from the pubic datasets~\cite{dolhansky2019-dfdc,rossler-2018-ff++} rather than generated by ourselves. In other words, we do not get access to the concrete approaches to manipulate the fake face.

\begin{figure*}[tb] 
\centering
\includegraphics[width=\linewidth]{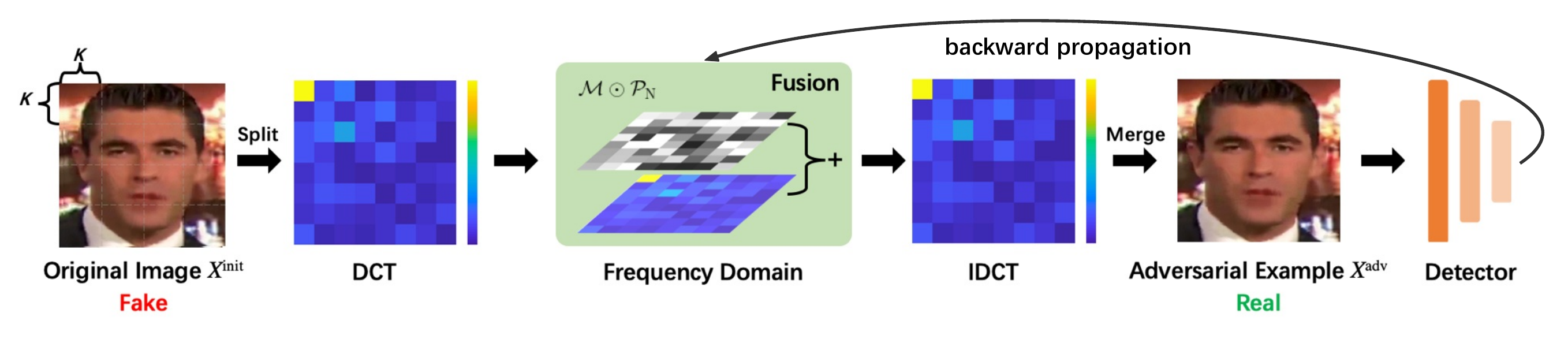}
\vspace{-0.30in}
\caption{The pipeline of frequency adversarial attack. We first split the input image into $K$$\times $$K$ blocks and apply DCT on each block to transfer them into the frequency domain. We then introduce the frequency perturbation $\mathcal{P}_N$ and a predefined weight matrix $\mathcal{M}$ that controls the step sizes in different frequency bands. After that, we implement IDCT and merge them into the adversarial example. In each iteration, we calculate the adversarial loss and update the perturbation $\mathcal{P}_N$ based on it.}
\label{fig:freatt}
\end{figure*}

\subsection{Face Forgery Detection}
Despite the creative applications of face forgery generation, this technology can be used abusively for malicious and unethical ways. Regarding its potential maleficence by the academic community, researchers attempt to detect if an image is manipulated or not to alleviate the danger, which is considered as a binary classification problem. Some works~\cite{afchar-wifs18-mesonet, rossler--iccv19-faceforensics++,  cozzolino-acmmm17-recasting, zhao-cvpr21-spa1} apply deep neural networks to extract discriminative features for face forgery detection. These methods only utilize the information from the spatial domain, which generally overfits the classification boundary. On the other hand, some works~\cite{liu-cvpr21-fre1,luo-cvpr21-fre2,durall2019unmasking, qian-eccv20-thinking,sun2021dual,wang-iconip20-face} observe the diversity of real faces and fake faces in the frequency domain and propose the face forgery detection method with the frequency clues. F\textsuperscript{3}-Net~\cite{qian-eccv20-thinking} integrates the frequency-aware decomposition and local frequency statistics into a whole learning framework to classify the real and fake faces. Luo et al.~\cite{luo-cvpr21-fre2} design several modules by taking full advantage of the high-frequency features at multiple scales to achieve higher accuracy. Furthermore, some methods~\cite{gu-aaai22-delving,masi-eccv20-two1,li-cvpr21-two2,chen-aaai21-local,gu-acmmm21-spatiotemporal,gu2021exploiting} integrate the spatial and frequency information into a whole framework to detect the fake face accurately. LRL~\cite{chen-aaai21-local} adopts a multi-task learning strategy with two output branches, where one branch is to learn the surface label and the other one aims to focus on the edge of the modified region. Li et al.~\cite{li-cvpr21-two2} combine the frequency clues with the spatial features to enlarge the difference between real faces and fake faces in the embedding space. Motivated by the diversity in the frequency domain, the proposed hybrid adversarial attack considers the effect on both domains to learn the robustness of existing face forgery detectors. For a complete comparison, we both select the spatial-based models and the frequency-based models to validate the effectiveness of the proposed attack method. 

\subsection{Adversarial Attack}
Different from face forgery generation, the aim of adversarial attack is to fool a machine rather than human beings. Generally, given a well-trained network, the goal of adversarial attack is to generate the adversarial examples that make the network predict wrongly. The category for adversarial attack can be divided into white-box attack~\cite{szegedy-iclr14-intriguing,goodfellow-2014-fgsm,madry-iclr18-pgd} and black-box attack~\cite{wang-cvpr21-delving, xie-cvpr19-difgsm,dong-cvpr19-tifgsm} roughly, which is based on the attacker gets access to the concrete structures and parameters of victim models or not. While the majority of existing attack methods focus on the multi classification task, adversarial attack has been investigated in many 
fields, such as object detection~\cite{xie-cvpr17-detection}, face recognition~\cite{yin-ijcai21-adv}, visual tracking~\cite{jia-eccv20-robust}, etc.

For adversarial attacks in face forgery detection, some works~\cite{li-cvpr21-exploring,neekhara-cvprw21-adversarial,carlini-cvprw20-evading,gandhi-ijcnn20-adversarial,hussain-wacv21-adversarial} explore the robustness of models in different settings.  Li et al.~\cite{li-cvpr21-exploring} manipulate the noise vectors and latent vectors of Style-GAN~\cite{wang-eccv16-stylegan} with gradients to fool the face forgery models. Neekhara et al.~\cite{neekhara-cvprw21-adversarial} perform adversarial attacks in a black-box setting for face forgery detection. Carlini et al.~\cite{carlini-cvprw20-evading} present the robustness of face forgery classifiers under various types of attack methods. The methods mentioned above generate the adversarial examples on the spatial domain, while some works~\cite{guo2018low,sharma-ijcai19-effectiveness} explore the frequency attack in other tasks. Since face forgery detection has a high relation with the frequency domain, we propose a novel attack method combined with the aspects of frequency domain to generate more imperceptible adversarial examples. 

\section{Method}
Let $X^{\text{init}}$ denote the original image, $f(X,\theta)$ denote the face forgery detector, and $y^{\text{gt}}$ denote the corresponding ground-truth label. 
Our aim is to generate the adversarial example $X^{\text{adv}}$ that makes the face forgery detector predict wrongly, i.e.,  $f(X^{\text{adv}},\theta)\neq y^{\text{gt}}$. During adversarial attack, the objective is to maximize the loss function $\mathcal{L} (X^{\text{adv}}, y^{\text{gt}})$, where $\mathcal{L} $ is the binary cross entropy loss in face forgery detection. The concrete optimization is defined as:
\begin{equation}\label{eq:advLoss1}
    \begin{aligned}
    {\rm arg\ max \ } \mathcal{L} (X^{\text{adv}}, y^{\text{gt}}), \ \ {\rm s.t.} \ {|| X^{\text{adv}} - X^{\text{init}}||}_{\rm p} < \epsilon,
    \end{aligned}
\end{equation}
where p is $l\textsubscript{p}$-norm to ensure the adversarial image close to the original image. We choose the untargeted attack to maximize the adversarial loss instead of the targeted attack due to the diversity of classification boundaries in different models. Although the targeted attack deteriorates the white-box model seriously, it has an extremely weak transferability to other models, which is prone to overfit on the specific network. 

\subsection{Spatial Adversarial Attack}
Existing attack methods are mostly considered as spatial adversarial attacks that modify the adversarial examples on the pixels. Due to the limited page, we only introduce two spatial adversarial attack methods that are utilized for comparisons in the experiments. More variants of these methods can refer to~\cite{dong-cvpr19-tifgsm}.
\vspace{-0.1in}

\paragraph{Fast Gradient Sign Method (FGSM).} FGSM~\cite{goodfellow-2014-fgsm} is a single-step attack method that calculates the perturbations based on the gradient of the adversarial loss. The optimization is defined as:
\begin{equation}\label{eq:fgsm}
X^{\text{adv}} = X^{\text{init}} + \epsilon \cdot {\rm sign} (\nabla_{X}  \mathcal{L} (X^{\text{adv}}, y^{\text{gt}}) ) .
\end{equation}
\paragraph{Projected Gradient Descent (PGD).} PGD ~\cite{madry-iclr18-pgd} is a multi-step variant of FGSM~\cite{xie-cvpr19-difgsm}. Meanwhile, it adopts a random initialization of perturbations at the first step. The update procedure is defined as:
\begin{equation}\label{eq:pgd}
\begin{aligned}
X_{0}^{\text{adv}}  = & X^{\text{init}},\\
X_{\text{n+1}}^{\text{adv}} =& {\rm Clip}\left \{X_{\text{n}}^{\text{adv}}+ \alpha \cdot {\rm sign} (\nabla_{X}  \mathcal{L}(X_{\text{n}}^{\text{adv}}, y^{\text{gt}})) \right \}.
\end{aligned}
\end{equation}

\subsection{Frequency Adversarial Attack}

\def\swone{0.3\linewidth}
\def\swthree{0.5\linewidth}
\begin{figure}[t]
    \small
	\begin{center}
	\includegraphics[width=0.9\linewidth]{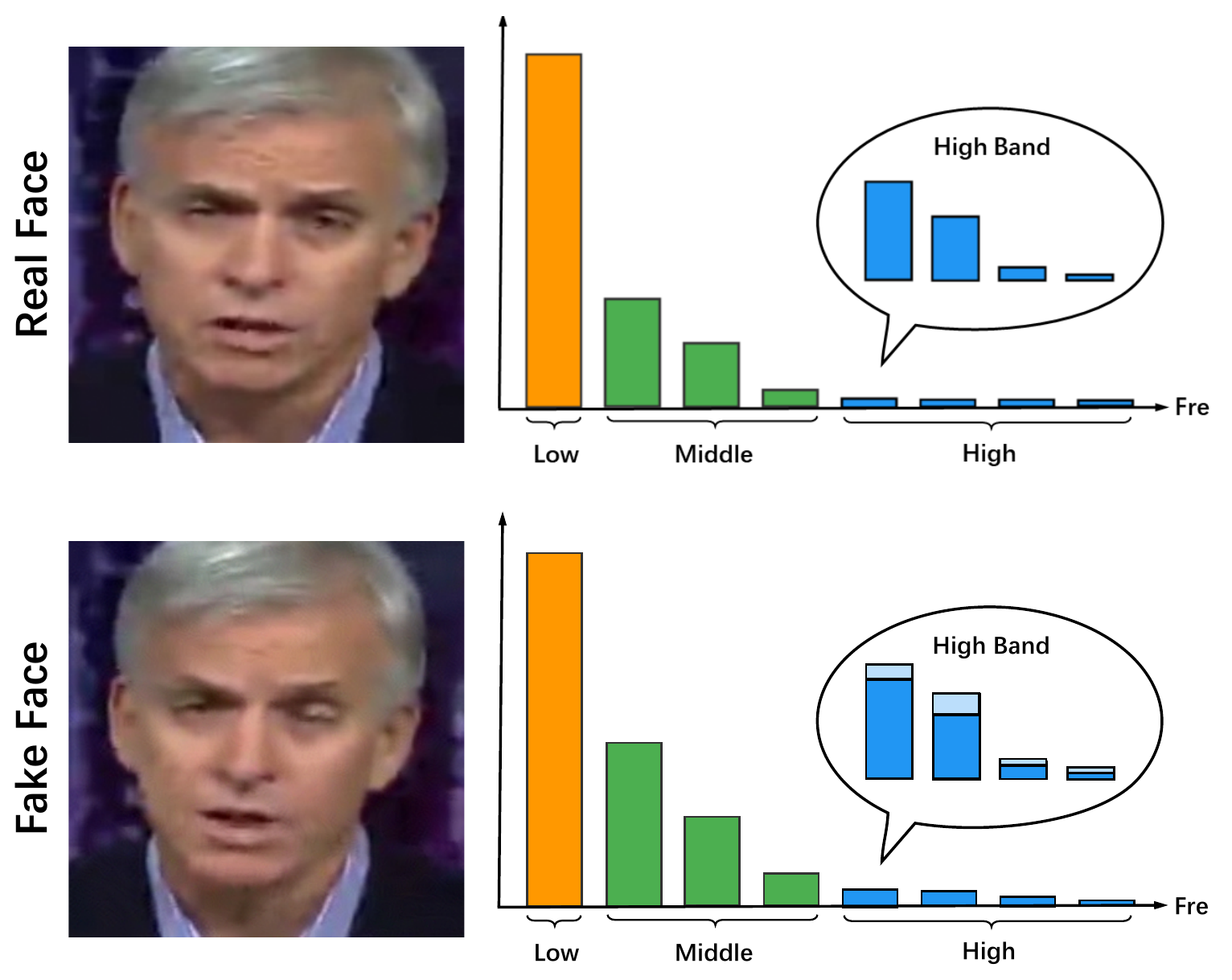}\\
	\vspace{-0.05in}
	(a) Image  \ \ \ \ \ \ \ \ \ \ \ \  (b)  Frequency distribution
	\end{center}
	\vspace{-0.2in}
\caption{The diversity between real faces and fake faces in the frequency domain. We select two examples with different labels from FaceForensics++~\cite{rossler-2018-ff++} and calculate the energy in different frequency bands. The energy of fake faces in the high frequency bands is richer than the one in real faces. }
\label{fig:difference}
\vspace{-0.05in}
\end{figure}

Previous studies~\cite{qian-eccv20-thinking,chen-aaai21-local} have proven the difference between the real face and the fake face in the frequency domain. Figure~\ref{fig:difference} demonstrates the diversity of energy in different frequency bands between a real face and a fake face. The low frequency region is related to the content of images accounting for most of the energy, while the high frequency region is related to the edge and texture information of images. The fake face shows more energy in the high frequency regions compared to the real one. Inspired by this observation, we propose a frequency adversarial attack to directly modify the energy in the frequency domain. Compared to the spatial attacks, our attack method hides the adversary in the frequency bands and decreases the redundant noise in the pixel level, leading to a more invisible attack. The pipeline of frequency adversarial attack is illustrated in Figure~\ref{fig:freatt}. We summarize the optimization procedure as follows:
\begin{equation}\label{eq:advLoss2}
    \begin{aligned}
    {\rm arg\ \ max \ \ } \mathcal{L}(\mathcal{D}^{\prime}(\mathcal{F} (\mathcal{D}(X^{\text{adv}}))), \theta, y^{\text{gt}}), \\
    \rm s.t. \ \ {|| \mathcal{D}(X^{\text{adv}})-\mathcal{D}(X^{\text{init}})||}_{p} < \epsilon,\ \ \ \ \ 
    \end{aligned}
\end{equation}
where $\mathcal{D}(\cdot)$ denotes discrete cosine transform (DCT), $\mathcal{D}^{\prime}(\cdot)$ denotes inverse discrete cosine transform (IDCT), $\mathcal{F}$ represents the fusion module to modify the energy in the frequency domain. Meanwhile, we utilize the $l_{\text{p}}$-norm to constrain the deviation of the original distribution of frequency. 

\begin{figure*}[tb] 
\centering
\includegraphics[width=\linewidth]{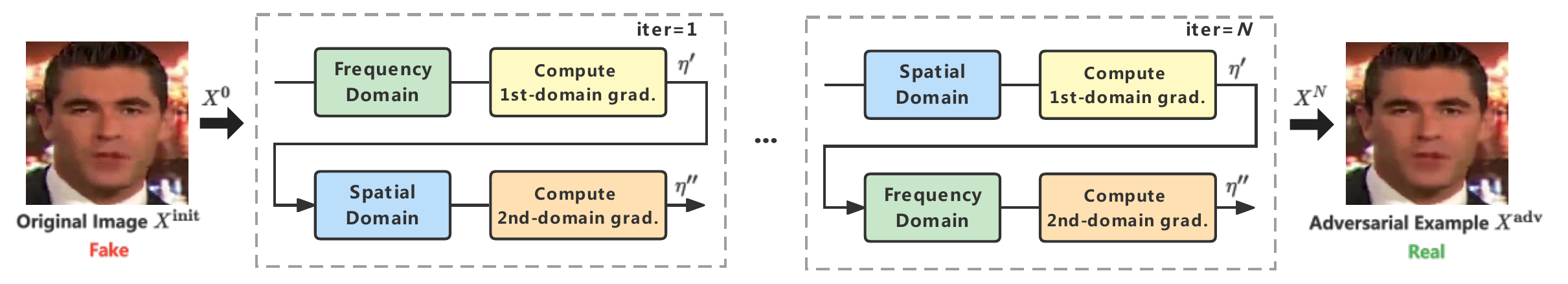}
\vspace{-0.30in}
\caption{The procedure of hybrid adversarial attack. To combine the adversarial attack in different domains, we calculate gradients from both domains in order and update the perturbations. Then, we switch the order of domains in the next step. After iterations, the adversarial example gathers the gradients from both domains, leading to a stronger adversarial attack on both white-box and black-box settings.}
\label{fig:hybrid}
\vspace{-0.1in}
\end{figure*}

\begin{algorithm}[t]
	\caption{Frequency Adversarial Attack}\label{algo:1}
	\KwIn{
		\hspace*{0.08in}Input image $X^{\text{init}}$, forensic detector $f(\cdot)$;}
	\KwOut{Adversarial examples $X^{\text{adv}}$; }
	Classify $f(X_0^{\text{adv}}, \theta)$ to get the true label $y^{\text{gt}}$ \; 
	Generate the initial perturbations $\mathcal{P}_0 \sim \mathcal{U}(0,1)$\;
	Initialize $X_0^{\text{adv}} = X^{\text{init}}$, $\hat{y} = y^{\text{gt}}$ \;
	\For{$n = 0$ \KwTo $N$}
	{
    	Split $X_{\text{n}}^{\text{adv}}$ into $K$$\times$$K$ blocks;\\
    	Apply the DCT on each block $\mathcal{D}(X_{\text{n}}^{\text{adv}})$\;
    	Calculate the adversarial loss $\mathcal{L}$ via Eq.~\ref{eq:advLoss2}\;
    	Update the perturbation $\mathcal{P}_{\text{n+1}}$ via Eq.~\ref{eq:perturb}\;
    	Fuse $\mathcal{P}_{\text{n+1}}$ and $\mathcal{M}$ into $\mathcal{D}(X_{\text{n}}^{\text{adv}})$ via Eq.~\ref{eq:fusion}\;
    	Apply the IDCT on each block $\mathcal{D}^{\prime}(\mathcal{F}(X^{\text{adv}}_n))$\;
    	Merge $K$$\times$$K$ blocks into $X_{\text{n+1}}^{\text{adv}}$\;
    	Classify $f(X_{\text{n+1}}^{\text{adv}}, \theta)$ to get the predicted result $\hat{y}$\;
	}
	\Return $X_{\text{n+1}}^{\text{adv}}$;
\end{algorithm}

For details, we first implement DCT to transfer the image from spatial domain to frequency domain by following ~\cite{qian-eccv20-thinking}. To balance the efficiency and quality of transformations, we split the original image into $K$$\times$$K$ blocks before DCT. For each block, we apply the DCT as follows:
\begin{equation}\label{eq:dct}
\begin{aligned}
\mathcal{D}(u,v)  = & \ \  c(u) \cdot c(v)
\sum_{i=0}^{N-1}\sum_{j=0}^{N-1}X(i,j) \\ &  \cos[{\frac{(2i+1)\pi}{2N}}u] \cdot\cos[{\frac{(2j+1)\pi}{2N}}v],\\
\end{aligned}
\end{equation}
where $X(i,j)$ is the value on the coordinate $(i,j)$ of image, $c(u)$ and $c(v)$ aim to make the DCT matrix orthogonal and $N$ is the size of each block. Then, we generate the initial perturbations $ \mathcal{P}$$\sim $$\mathcal{U}(0,1)$ to inject on the frequency band. When the RGB image transfers into the frequency domain, the range of energy in different frequencies are lopsided as shown in Figure~\ref{fig:difference}. Therefore, we propose a matrix $\mathcal{M}$ with adaptable step sizes, which is based on the proportion of each frequency band to balance the influence of lopsided energy. Moreover, the matrix $\mathcal{M}$ is dynamically reset for diverse inputs to maintain the visual quality. The complete fusion module is defined as: 
\begin{equation}\label{eq:fusion}
\mathcal{F}(X^{\text{adv}}_{\text{n}}) = \mathcal{D}(X^{\text{adv}}_{\text{n}}) + \mathcal{M} \odot \mathcal{P}_{\text{n+1}}, 
\end{equation}
where $\odot$ is Hadamard product. During the optimization, $ \mathcal{P}_{\text{n+1}}$ is updated as follows:
\begin{equation}\label{eq:perturb}
 \mathcal{P}_{\text{n+1}} = \mathcal{P}_{\text{n}} + \lambda \cdot {\rm sign} (\nabla_{\mathcal{P}}  \mathcal{L}(\mathcal{D}^{\prime}(\mathcal{F} (\mathcal{D}(X_{\text{n}}^{\text{adv}}))), \theta, y^{\text{gt}})),
\end{equation}
where $\lambda$ is the step size in each iteration. After that, we apply IDCT to transfer each block in the frequency domain back to the spatial domain. Note that the solo operation of DCT and IDCT is non-destructive, where no block artifacts are introduced in the process. When it reaches the maximum iteration or classifies the $X^{\text{adv}}$ as a wrong label, we end up the loop and output the final adversarial example $X^{\text{adv}}$. The pseudo code is shown in Algorithm~\ref{algo:1}.

\subsection{Hybrid Adversarial Attack}
Meta-learning~\cite{finn-icml17-meta} aims to train a model that can quickly adapt to a new task with only a few training steps and training data, which is summarized as  `learn to learn'. Inspired by the idea of meta-learning~\cite{nichol2018first}, we propose a hybrid adversarial attack combined with the spatial domain and the frequency domain. Different from the vanilla meta-learning that updates a new model through the training data, we directly utilize the gradient from different domains to iteratively update the adversarial perturbations. Our hybrid adversarial attack can gather the virtues from both domains and integrate them in a compatible way. The hybrid perturbations improve the effectiveness of adversarial attacks in white-box settings and have a strong transferability on other models. The procedure of our hybrid adversarial attack is shown in Figure~\ref{fig:hybrid}. 

Let $\mathcal{A}_S$ and $\mathcal{A}_F$ denote the adversarial attack in spatial and frequency domain, respectively. At first, we compute the gradient based on the adversarial loss in the frequency domain. The optimization of $\mathcal{A}_F$ in the frequency domain is calculated by:
\begin{equation}\label{eq:fre}
 \eta^{\prime} = \eta - \gamma_{f} \cdot \nabla_{\eta} \mathcal{L}_{\mathcal{A}_F} (\eta, \theta, y^{\text{gt}}),
\end{equation}
where $\eta^{\prime}$ is frequency values and $\gamma_{f}$ is the step size in the frequency domain. Then, we compute the gradient based on the adversarial loss in the spatial domain. The process of spatial attack $\mathcal{A}_S$ is formulated as:
\begin{equation}\label{eq:spa}
 \eta^{\prime\prime} =  \eta^{\prime}- \gamma_{s} \cdot \nabla_{\eta^{\prime}} \mathcal{L}_{\mathcal{A}_S} (\eta^{\prime}, \theta, y^{\text{gt}}),
\end{equation}
where $ \eta^{\prime\prime}$ is pixel values and $\gamma_{s}$ is the step size in the spatial domain. The detailed procedure of adversarial attacks in each domain follows the above sections. We select PGD~\cite{madry-iclr18-pgd} as our spatial attack and the proposed frequency attack method as our frequency attack, respectively. Note that we remove the clip function in both attacks and add it at the end of each iteration. After each iteration, we switch the order of frequency attack $\mathcal{A}_F$ and spatial attack $\mathcal{A}_S$, and repeat the whole procedure. The complete algorithm of our hybrid adversarial attack is presented in Algorithm~\ref{algo:2}.

\begin{algorithm}[t]
	\caption{Hybrid Adversarial Attack}\label{algo:2}
	\KwIn{
		\hspace*{0.08in}Input image $X^{\text{init}}$,  forensic detector $f(\cdot)$,\\
		\hspace*{0.54in}spatial attack $\mathcal{A}_S$, frequency attack $\mathcal{A}_F$;}
	\KwOut{Adversarial examples $X^{\text{adv}}$; }
	Classify $f(X_0^{\text{adv}}, \theta)$ to get the true label $y^{\text{gt}}$ \; 
	Initialize $X_0^{\text{adv}} = X^{\text{init}}$, $\hat{y} = y^{\text{gt}}$ \;
	\For{$n = 0$ \KwTo $N$}
	{
    	Calculate the adversarial loss $\mathcal{L}_{\mathcal{A}_F}$ via Eq.~\ref{eq:advLoss2}\;
    	Update the perturbation $\eta^{\prime}$ via Eq.~\ref{eq:fre}\;
    	Calculate the adversarial loss $\mathcal{L}_{\mathcal{A}_S}$ via Eq.~\ref{eq:advLoss1}\;
        Update the perturbation $\eta^{\prime\prime}$ via Eq.~\ref{eq:spa}\;
    	Clip the output images $X_{\text{n+1}}^{\text{adv}}$ with the $l_{\text{p}}$-norm;\\
    	Classify $f(X_{\text{n+1}}^{\text{adv}}, \theta)$ to get the predicted result $\hat{y}$\;
    	Switch the order of attacks $\mathcal{A}_S$ and $\mathcal{A}_F$;\\
	}
	\Return $X_{\text{n+1}}^{\text{adv}}$;
\end{algorithm}

\begin{table*}[tb]
\small
\centering
\caption{The accuracy of spatial-based and frequency-based face forgery detectors on the DFDC~\cite{dolhansky2019-dfdc}  and  FaceForensics++~\cite{rossler--iccv19-faceforensics++} datasets.}
\label{table:orig}
\vspace{-0.05in}
\begin{tabular} {|l|ccc|cc|}
\hline
Dataset & EfficientNet\_b4~\cite{tan2019efficientnet} & ResNet\_50~\cite{resnet} & XceptionNet~\cite{xception} \ \ \ & \ \ \ F\textsuperscript{3}-Net~\cite{chen-aaai21-local}\ \ \  &\ \ \ LRL~\cite{qian-eccv20-thinking}\ \ \ \\
\hline\hline
DFDC~\cite{dolhansky2019-dfdc} & 91.1\% & 78.7\% & 88.0\% & 69.8\% & 90.4\% \\
FaceForensics++~\cite{rossler-2018-ff++} & 94.3\% & 89.1\% & 92.7\% & 88.8\% & 98.2\% \\
\hline
\end{tabular}
\vspace{-0.1in}
\end{table*}

\section{Experiment}
In this section, we first introduce the experimental setup. Then, we evaluate the performance of the proposed attack method with the single attacks and the ensemble attacks on the spatial-based models. We further validate our attack method on the frequency-based models. In addition, we conduct ablation studies on the variations of our method and different frequency bands. We finally evaluate the image quality of our method qualitatively and quantitatively.

\subsection{Experimental Setup}
\paragraph{Datasets.} DFDC~\cite{dolhansky2019-dfdc} is a challenging dataset with a variety of anonymous manipulations and perturbations. We randomly select 1000 fake face images from the DFDC dataset. FaceForensics++~\cite{rossler-2018-ff++} is a popular dataset containing real videos from YouTube and corresponding fake videos, consisting of Deepfake~\cite{deepfake}, Face2Face~\cite{thies-cvpr16-face2face}, FaceSwap~\cite{facesawp}  and NeuralTextures~\cite{thies-tog19-neural}. We totally choose 560 (140$\times$4) individual frames from each fake face video.
\vspace{-0.15in}

\paragraph{Models.} For the spatial-based face forgery detectors, we choose three spatial-based classification networks, i.e., EfficientNet\_b4~\cite{tan2019efficientnet}, ResNet\_50~\cite{resnet}, and XceptionNet~\cite{xception}. For the frequency-based models, we consider the state-of-the-art face forgery detectors, i.e., F\textsuperscript{3}-Net~\cite{qian-eccv20-thinking} and LRL~\cite{chen-aaai21-local}. All these models are trained by following the corresponding papers. The accuracy of these models on the selected images from different datasets are summarized in Table~\ref{table:orig}.
\vspace{-0.15in}

\begin{table}[tb]
\small
\centering
\caption{The attack success rate of fake faces on spatial-based models on the DFDC~\cite{dolhansky2019-dfdc} dataset.}
\vspace{-0.1in}
\label{table:single dfdc}
\begin{tabular*} {8.3cm} {@{\extracolsep{\fill}}|l|l|ccc|}
\hline
Model  & Attack    & Eff\_b4~\cite{tan2019efficientnet} & Res50~\cite{resnet} & Xcep~\cite{xception} \\
\hline\hline
\multirow{3}{*}{Eff\_b4~\cite{tan2019efficientnet}}  
& FGSM   & 33.2\% & ~7.1\% & ~2.3\%   \\ 
& PGD    & 77.7\% & ~8.7\% & ~1.8\%   \\ 
& Ours   & \textbf{97.1}\% & \textbf{20.1}\% &\textbf{~2.7}\%   \\ 
\hline \hline
\multirow{3}{*}{Res50~\cite{resnet}}  
& FGSM   & ~0.0\% & 36.7\% & ~0.9\%   \\ 
& PGD    & ~0.0\% & 85.4\% & ~0.0\%   \\  
& Ours   & \textbf{23.2}\% & \textbf{87.8}\% & \textbf{24.1}\%   \\ 
\hline \hline
\multirow{3}{*}{Xcep~\cite{xception}}     
& FGSM   & 0.0\% & 8.4\% & 45.6\%   \\ 
& PGD    & 0.0\% & 10.1\% & 72.3\%   \\ 
& Ours   & \textbf{~1.2}\% & \textbf{14.3}\% & \textbf{77.5}\%   \\ 
\hline
\end{tabular*}
\vspace{-1ex}
\end{table}

\paragraph{Evaluation metrics.} For DFDC and FaceForensics++, we both choose the attack success rate as the evaluation metric. It is defined as the proportion of successfully attack images in all images that are classified as fake faces, i.e., $\frac{1}{N}\sum_{n=1}^{N} f(X^{\text{adv}},\theta) \neq f(X^{\text{init}},\theta)$. For the image quality assessment, we utilize MSE, PSNR and SSIM as the evaluation metrics to present the difference between the generated adversarial example and the original image. 
\vspace{-0.15in}

\paragraph{Implementation details.} The input size of images for three spatial-based models is 320$\times$320$\times$3. And the input sizes for F\textsuperscript{3}-Net~\cite{qian-eccv20-thinking} and LRL~\cite{chen-aaai21-local} are 299$\times$299$\times$3 and 320$\times$320$\times$3, respectively. We resize the adversarial examples to the corresponding size for the transfer attack. As for the parameters for attacks, we set the maximum perturbation of each pixel to be $\epsilon=0.1$ for both FGSM and PGD. We also use PGD as the spatial attack for the hybrid  attack.

\subsection{Attack on Spatial-based Models}
We compare the proposed method with FGSM~\cite{goodfellow-2014-fgsm} and PGD~\cite{madry-iclr18-pgd} on attacking spatial-based models. Table~\ref{table:single dfdc} and Table~\ref{table:single ff++} report the attack success rates on the DFDC ~\cite{dolhansky2019-dfdc} and FaceForensics++~\cite{rossler-2018-ff++} datasets, respectively. We consider the basic classifiers in the first column to generate the adversarial examples and transfer them on the other networks to evaluate. The diagonal blocks indicate white-box attacks, while the off-diagonal blocks indicate their transferablility as black-box attacks. As Table~\ref{table:single dfdc} and Table~\ref{table:single ff++} report, the proposed method outperforms FGSM and PGD for the white-box attack and gains higher attack success rates for the black-box attack. For instance, the adversarial examples generated by Res50 with our method get success rates of 41.4\% on Eff\_b4  and 49.6\% on Xcep on the FaceForensics++~\cite{rossler-2018-ff++} dataset, which is 38.2\% higher than FGSM and 37.5\% higher than PGD on Eff\_b4, and 47.5\% greater than FGSM and 47.3\% higher than PGD on Xcep. It suggests that the proposed method combined with the frequency attack enhances the transferability of adversarial examples. Due to the obvious diversity of structure between Eff\_b4 and Xcep, the adversarial attacks between two networks have limited transferablility on each other.

\begin{table}[tb]
\small
\centering
\caption{The attack success rate of fake faces on spatial-based models on the FaceForensics++~\cite{rossler-2018-ff++} dataset.}\label{table:single ff++}
\vspace{-0.1in}
\begin{tabular*} {8.3cm} {@{\extracolsep{\fill}}|l|l|ccc|}
\hline
Model  & Attack    & Eff\_b4~\cite{tan2019efficientnet} & Res50~\cite{resnet} & Xcep~\cite{xception} \\
\hline\hline
\multirow{3}{*}{Eff\_b4~\cite{tan2019efficientnet}}  
& FGSM   & 38.7\% &~4.8\% & ~0.9\%   \\ 
& PGD    & 71.6\% & ~1.3\% & ~0.3\%   \\ 
& Ours   & \textbf{83.2}\% & \textbf{22.7}\% & \textbf{~1.4}\%   \\ 
\hline \hline
 \multirow{3}{*}{Res50~\cite{resnet}}  
& FGSM   & ~3.2\% & 32.0\% & ~2.1\%   \\ 
& PGD    & ~3.9\% & 60.2\% & ~2.3\%   \\  
& Ours   & \textbf{41.4}\% & \textbf{65.4}\% & \textbf{49.6}\%   \\ 
\hline \hline
\multirow{3}{*}{Xcep~\cite{xception}}     
& FGSM   & ~1.1\% & ~4.1\% & 18.9\%   \\ 
& PGD    & ~1.1\% & ~7.7\% & 61.6\%   \\ 
& Ours   & \textbf{~1.5}\% & \textbf{~8.5}\% & \textbf{70.5}\%   \\ 
\hline
\end{tabular*}
\vspace{-1ex}
\end{table}

\subsection{Ensemble Attack on Spatial-based  Models}
As stated in~\cite{xie-cvpr19-difgsm}, the adversarial examples with an ensemble of multiple networks achieve much stronger attack performance. We utilize an ensemble of two networks to attack the other one in three ways: ensemble in pixel, ensemble in loss, and ensemble in logits. The attack results on two datasets are summarized in Table~\ref{table:ensemble dfdc} and Table~\ref{table:ensemble ff++}, respectively. We consider all three networks and the sign  `-' in the first column indicates the network not used during attacks. Thus, the diagonal blocks indicate transfer attacks (i.e., black-box setting) and the off-diagonal blocks indicate the white-box attacks. From both datasets, we observe that the ensemble in logits performs the strongest attack performance in most cases. In the DFDC~\cite{dolhansky2019-dfdc} dataset, when attacking on Res50 network, the ensemble in logits of Res50 and Eff\_b4 obtains a 7.8\% higher than the single network Res50 under the white-box setting. Besides, for the FaceForensics++~\cite{rossler-2018-ff++} dataset, the ensemble in logits of Res50 and Eff\_b4 achieves a 38.2\% success rate on Xcep, while the single network Res50 only gets an 8.5\% success rate under the black-box setting. To sum up, an ensemble of different models can increase the diversity of structures, leading to a greater transferability to other models.

\begin{table}[tb]
\small
\centering
\caption{The attack success rate of fake faces with ensemble attacks on the DFDC~\cite{dolhansky2019-dfdc} dataset.}\label{table:ensemble dfdc}
\vspace{-0.1in}
\begin{tabular*} {8.3cm} {@{\extracolsep{\fill}}|l|l|ccc|}
\hline
Model  & Ens.  & Eff\_b4~\cite{tan2019efficientnet} & Res50~\cite{resnet} & Xcep~\cite{xception} \\
\hline\hline
\multirow{3}{*}{-Eff\_b4~\cite{tan2019efficientnet}}    
& Pixel   & \textbf{~4.1}\% & 86.9\% & 44.4\%   \\ 
& Loss   & ~2.3\% & 72.0\% & 57.3\%   \\ 
& Logit  & ~2.4\% & \textbf{88.5}\% & \textbf{77.5}\%   \\ 
\hline \hline
\multirow{3}{*}{-Res50~\cite{resnet}}   
& Pixel    & 79.7\% & 20.4\% & 28.5\%   \\ 
& Loss    & 71.1\% & 18.1\% & 46.2\%   \\ 
& Logit   & \textbf{93.1}\% & \textbf{22.1}\% & \textbf{65.3}\%   \\  
\hline \hline
\multirow{3}{*}{-Xcep~\cite{xception}} 
& Pixel  & 72.4\% & 86.2\% & 13.1\%   \\  
& Loss  & 76.4\% & 75.8\% & \textbf{14.7}\%   \\ 
& Logit  & \textbf{95.4}\% & \textbf{95.6}\% & 12.7\%   \\ 
\hline
\end{tabular*}
\end{table}

\begin{table}[tb]
\small
\centering
\caption{The attack success rate of fake faces with ensemble attacks on the FaceForensics++~\cite{rossler-2018-ff++} dataset.}\label{table:ensemble ff++}
\vspace{-0.1in}
\begin{tabular*} {8.3cm} {@{\extracolsep{\fill}}|l|l|ccc|}
\hline
Model  & Ens.  & Eff\_b4~\cite{tan2019efficientnet} & Res50~\cite{resnet} & Xcep~\cite{xception} \\
\hline\hline
\multirow{3}{*}{-Eff\_b4~\cite{tan2019efficientnet}}    
& Pixel   & 23.1\% & 64.0\% & 71.1\%   \\ 
& Loss   & 27.0\% & 52.9\% & \textbf{79.1}\%   \\ 
& Logit  & \textbf{27.5}\% & \textbf{69.5}\% & 68.6\%   \\ 
\hline \hline
\multirow{3}{*}{-Res50~\cite{resnet}}   
& Pixel   & 77.3\% & 26.6\% & 49.1\%   \\ 
& Loss   & 55.2\% & 21.8\% & 57.9\%   \\ 
& Logit  & \textbf{78.0}\% & \textbf{29.8}\% & \textbf{76.1}\%   \\  
\hline \hline
\multirow{3}{*}{-Xcep~\cite{xception}} 
& Pixel   & 78.9\% & \textbf{64.6}\% & 30.7\%   \\  
& Loss   & \textbf{83.4}\% & 57.3\% & 36.2\%   \\ 
& Logit  & 82.0\% & 64.4\% & \textbf{38.2}\%   \\ 
\hline
\end{tabular*}
\vspace{-1ex}
\end{table}

\subsection{Attack on Frequency-based Models}

The proposed hybrid attack is related to the frequency domain. To further illustrate its effectiveness, we also select two frequency-based face forgery detection methods, i.e., F\textsuperscript{3}-Net~\cite{chen-aaai21-local}  and LRL~\cite{qian-eccv20-thinking}. Both methods collect the frequency information to distinguish the diversity between the real and fake faces to detect. Table~\ref{table:fre dfdc} and Table~\ref{table:fre ff++} report the attack results on the DFDC~\cite{dolhansky2019-dfdc} and FaceForensics++~\cite{rossler-2018-ff++} datasets, respectively. For better comparison, we also conduct the experiments of two detectors with FGSM~\cite{goodfellow-2014-fgsm} and PGD~\cite{madry-iclr18-pgd}. Moreover, we test the transferability of adversarial examples that our hybrid attack generates when attacking the spatial-based detectors. Briefly, our hybrid adversarial attack associated with the frequency domains achieves favorable white-box attacks in both datasets, where $\sim$90\% of fake images are classified wrongly as real faces. For the transfer attack, our method is marginally greater than the spatial attacks. When using the spatial-based detectors for transfer attacks, Res50 outperforms the other two networks with success attack rates of 12.8\% for F\textsuperscript{3}-Net and 43.6\% for LRL on DFDC~\cite{dolhansky2019-dfdc}, and 7.1\% for F\textsuperscript{3}-Net and 57.5\% for LRL on FaceForensics++~\cite{rossler-2018-ff++}. The proposed hybrid attack with the frequency domain strengthens the transferability of networks on the frequency-based models as well. 

\begin{table}[tb]
\small
\centering
\caption{The attack success rate of fake faces on frequency-based models on the DFDC~\cite{dolhansky2019-dfdc} dataset.}\label{table:fre dfdc}
\vspace{-0.1in}
\begin{tabular} {|l|l|cc|}
\hline
Model \ \ \ \ \ \ \ \ & Attack \ \ \ \ \  & \ \ \ \ F\textsuperscript{3}-Net~\cite{chen-aaai21-local} \ \ \  \ & \ \ \ \  LRL~\cite{qian-eccv20-thinking} \ \ \ \  \\
\hline\hline
\multirow{3}{*}{F\textsuperscript{3}-Net~\cite{chen-aaai21-local}}  
& FGSM  & 43.5\% & \ \ 9.6\%   \\ 
& PGD     & 97.6\% & \ \ 4.0\%   \\ 
& Ours    & \textbf{98.7}\% & \textbf{10.3}\%   \\ 
\hline \hline
\multirow{3}{*}{LRL~\cite{qian-eccv20-thinking}}    
& FGSM      & ~2.3\% & 71.3\%   \\
& PGD   & ~3.0\% & \textbf{100.0}\% \ \   \\  
& Ours  & \textbf{~5.5}\% & \textbf{100.0}\% \ \   \\ 
\hline \hline
Eff\_b4~\cite{tan2019efficientnet}  & Ours  & ~7.4\% & ~8.5\%   \\ 
Res50~\cite{resnet}                          & Ours  & \textbf{12.8}\% & \textbf{43.6}\%   \\ 
Xcep~\cite{xception}                        & Ours  &~7.6\% & ~9.1\%   \\  
\hline
\end{tabular}
\end{table}

\begin{table}[tb]
\small
\centering
\caption{The attack success rate of fake faces on frequency-based models on the FaceForensics++~\cite{rossler-2018-ff++} dataset.}\label{table:fre ff++}
\vspace{-0.1in}
\begin{tabular} {|l|l|cc|}
\hline
Model \ \ \ \ \ \ \ \ & Attack \ \ \ \ \  & \ \ \ \ F\textsuperscript{3}-Net~\cite{chen-aaai21-local} \ \ \  \ & \ \ \ \  LRL~\cite{qian-eccv20-thinking} \ \ \ \  \\
\hline\hline
\multirow{3}{*}{F\textsuperscript{3}-Net~\cite{chen-aaai21-local}}  
& FGSM      & 24.8\% & ~7.7\%   \\ 
& PGD    & 80.9\% & 28.7\%   \\ 
& Ours   & \textbf{82.5}\% & \textbf{36.2}\%   \\ 
\hline \hline
\multirow{3}{*}{LRL~\cite{qian-eccv20-thinking}}    
& FGSM      & ~0.2\% & 68.6\%   \\ 
& PGD    & ~0.0\% & 98.7\%   \\  
& Ours   & \textbf{~0.5}\% & \textbf{99.3}\%   \\ 
\hline \hline
Eff\_b4~\cite{tan2019efficientnet}        & Ours  & ~0.5\% & 11.8\%   \\ 
Res50~\cite{resnet}                     & Ours  & \textbf{~7.1}\% & \textbf{57.5}\%   \\ 
Xcep~\cite{xception}                    & Ours  & ~1.1\% & 19.5\%   \\  
\hline
\end{tabular}
\vspace{-1ex}
\end{table}

\subsection{Ablation Study}
We conduct a series of ablation studies on the proposed attack method. Due to the limited page, we only use Res50~\cite{resnet} as the threat model to generate adversarial examples on the FaceForensics++~\cite{rossler-2018-ff++} dataset and show its transferability on Eff\_b4~\cite{tan2019efficientnet} and Xcep~\cite{xception}. 
\vspace{-0.15in}

\paragraph{Variants of our method.} We perform some variants of our method to analyze the effects on different domains. Concretely, we first only apply the spatial attack to craft adversarial examples. Then, we only implement the frequency attack. Besides, we simply sum the perturbations from the spatial and frequency domain. The concrete results are summarized in Table~\ref{table:ab1}. While the sole spatial attack is prone to overfit the existing model and yields a limited transferability, the sole frequency attack always falls into local optimization and retains a fixed loss for binary classification, leading to a limited attack ability. For the combination of perturbations, the simple sum of perturbations from two domains improves the performance of white-box attack but weakens its transferability. Our hybrid adversarial attack has more aggressive attack performance on both white-box and black-box attacks, which indicates that our method maintains both benefits from each domain and integrates them in a compatible way.
\vspace{-0.15in}

\paragraph{Frequency bands.} To study the effect of different frequency bands, we divide the whole frequency band into three bands, i.e., low band, middle band, and high band, and only attack one of the bands with the hybrid adversarial attack. Table~\ref{table:ab2} reports the attack performance for different frequency bands. Compared with middle bands and high bands, the hybrid adversarial attack on low bands performs favorably under both white-box attacks (e.g., Res50~\cite{resnet}) and black-box attacks (e.g., Eff\_b4~\cite{tan2019efficientnet} and Xcep~\cite{xception}), since the low bands carry more content information of images and generate more perturbations in the spatial domain. 

\begin{table}[tb]
\small
\centering
\caption{The attack success rate of fake faces with variants of our method on the FaceForensics++~\cite{rossler-2018-ff++} dataset.}\label{table:ab1}
\vspace{-0.1in}
\begin{tabular*} {8.3cm} {@{\extracolsep{\fill}}|l|ccc|}
\hline
Method    & Eff\_b4~\cite{tan2019efficientnet} & Res50~\cite{resnet} & Xcep~\cite{xception} \\
\hline\hline
Spatial attack    & ~3.9\% & 60.2\% & ~2.3\%   \\ 
Frequency attack  & 14.5\% & 29.3\% & 15.9\%   \\ 
Sum attack        & ~7.7\% & 61.2\% & 12.1\%   \\
\hline \hline
Hybrid attack     & \textbf{41.4}\% & \textbf{65.4}\% & \textbf{49.6}\%   \\ 
\hline
\end{tabular*}
\vspace{-1ex}
\end{table}

\begin{table}[tb]
\small
\centering
\caption{The attack success rate of fake faces for attacking different frequency bands on the FaceForensics++~\cite{rossler-2018-ff++} dataset.}\label{table:ab2}
\vspace{-0.1in}
\begin{tabular*} {8.3cm} {@{\extracolsep{\fill}}|l|ccc|}
\hline
Frequency \ \   & Eff\_b4~\cite{tan2019efficientnet} & Res50~\cite{resnet} & Xcep~\cite{xception} \\
\hline\hline
Low bands       & 39.5\% & 65.2\% & 49.1\%   \\ 
Middle bands   & 38.4\% & 62.7\% & 48.0\%   \\ 
High bands      & 36.2\% & 61.2\% & 47.1\%   \\
\hline \hline
All bands    & \textbf{41.4}\% & \textbf{65.4}\% & \textbf{49.6}\%   \\ 
\hline
\end{tabular*}
\vspace{-1ex}
\end{table}

\begin{table}[tb]
\small
\centering
\caption{Quantitative evaluation of adversarial examples generated by FGSM~\cite{goodfellow-2014-fgsm}, PGD~\cite{madry-iclr18-pgd} and our method on the FaceForensics++~\cite{rossler-2018-ff++} dataset.}\label{table:qua}
\vspace{-0.1in}
\begin{tabular*} {8.3cm} {@{\extracolsep{\fill}}|l|ccc|}
\hline
Attack method\ \   & MSE ($\downarrow$)& PSNR ($\uparrow$) & SSIM ($\uparrow$)\\ 
\hline\hline
FGSM      & 0.0279 & 23.3 & 0.0881   \\ 
FGD         & 0.0238 & 30.4 & 0.1343   \\ 
\hline \hline
Ours     & \textbf{0.0027} &  \textbf{42.7} &  \textbf{0.1763}   \\ 
\hline
\end{tabular*}
\vspace{-0.1in}
\end{table}

\subsection{Image Quality Assessment}
In order to illustrate the superior image quality by our method, we analyze the generated adversarial examples qualitatively and quantitatively. In Figure~\ref{fig:more}, we visualize adversarial examples crafted by FGSM~\cite{goodfellow-2014-fgsm}, PGD~\cite{madry-iclr18-pgd} and our method. The adversarial examples by FGSM and PGD have obvious noise patterns when zooming in, while the ones generated by our method are more imperceptible to observers. In addition, we use the common metrics for image quality assessment to calculate the difference to the original images. Table~\ref{table:qua} reports the quantitative results of MSE, PSNR and SSIM, where the image quality of our method outperforms other attacks by a large margin. It suggests that the proposed hybrid attack has a strong attack ability and maintains the image quality highly.

\def\swfive{0.24\linewidth}
\begin{figure}[t]
    \small
	\begin{center}
	\includegraphics[width=\swfive]{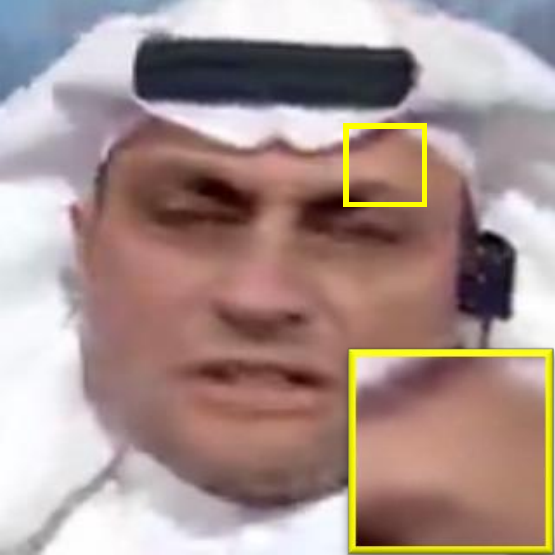}
	\includegraphics[width=\swfive]{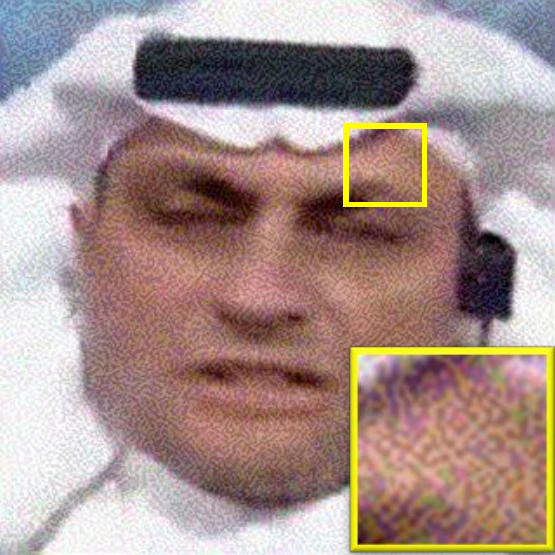}
	\includegraphics[width=\swfive]{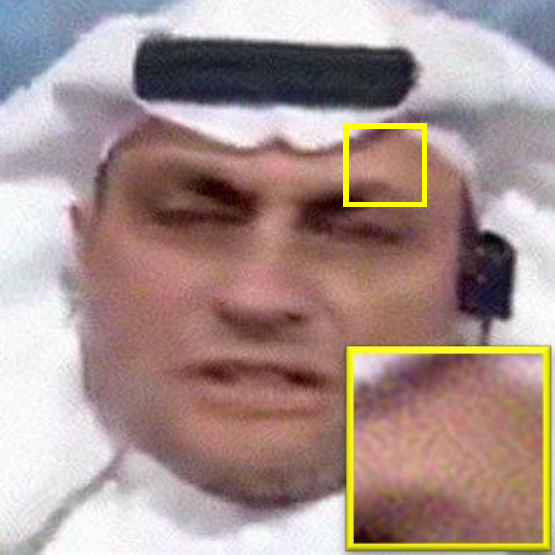}
	\includegraphics[width=\swfive]{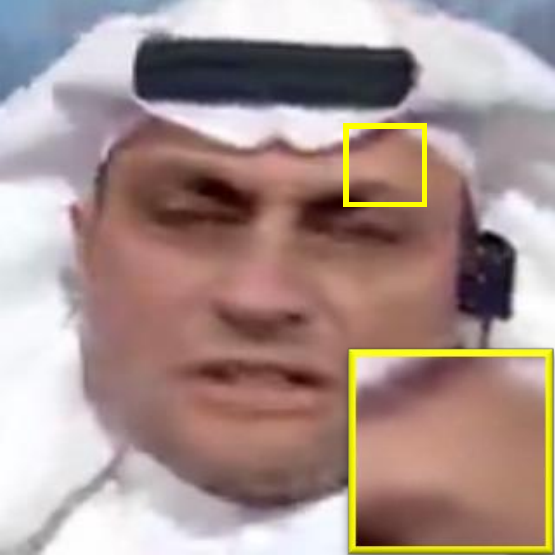}\\
	\includegraphics[width=\swfive]{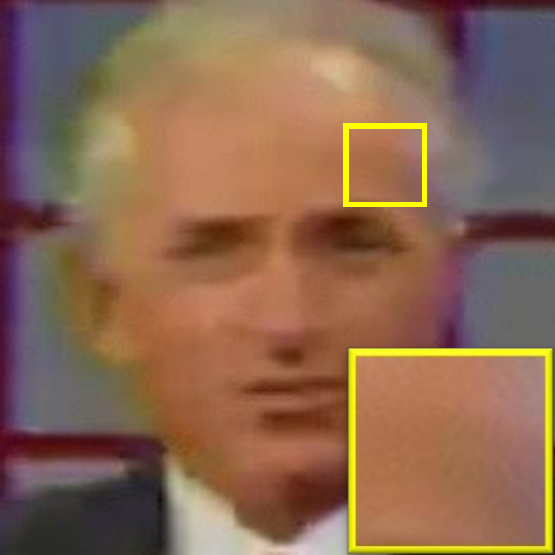}
	\includegraphics[width=\swfive]{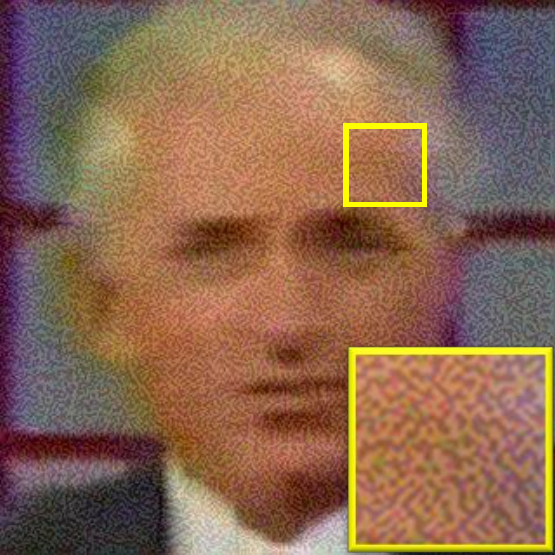}
	\includegraphics[width=\swfive]{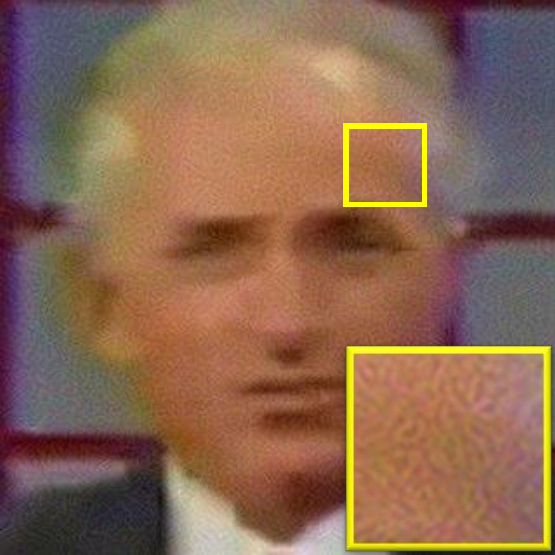}
	\includegraphics[width=\swfive]{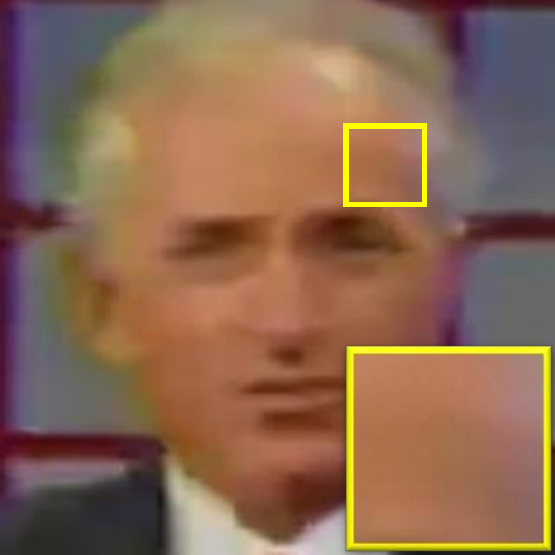}\\
	\includegraphics[width=\swfive]{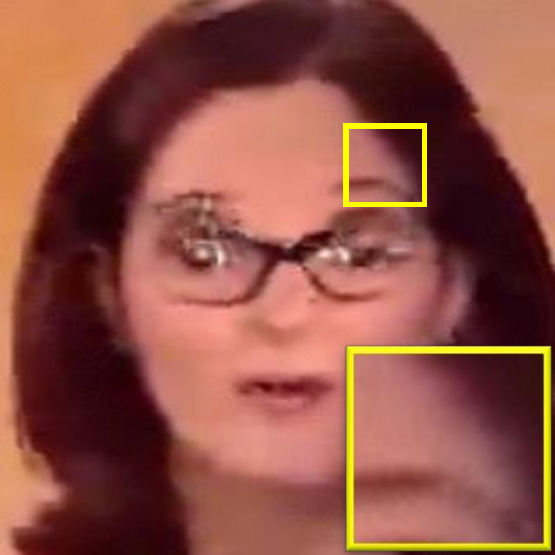}
	\includegraphics[width=\swfive]{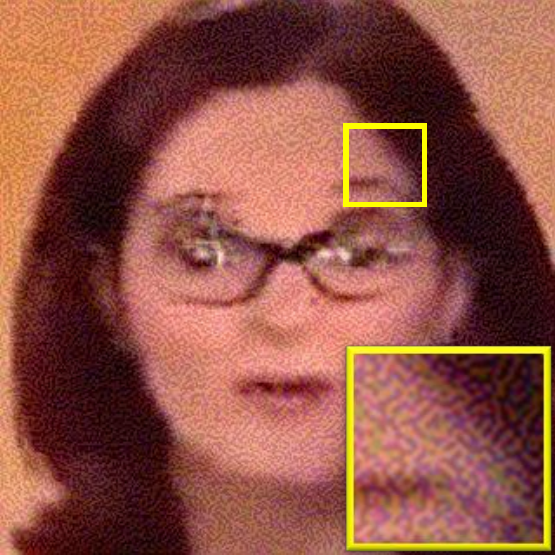}
	\includegraphics[width=\swfive]{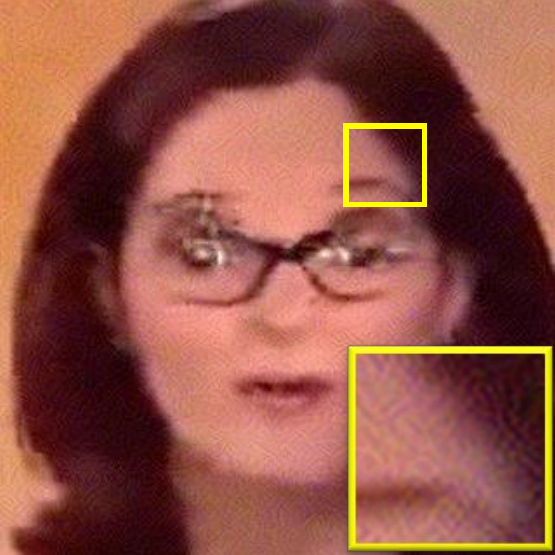}
	\includegraphics[width=\swfive]{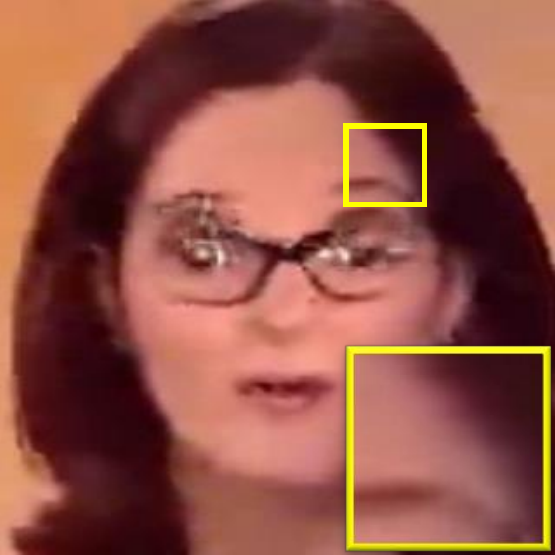}\\
	\includegraphics[width=\swfive]{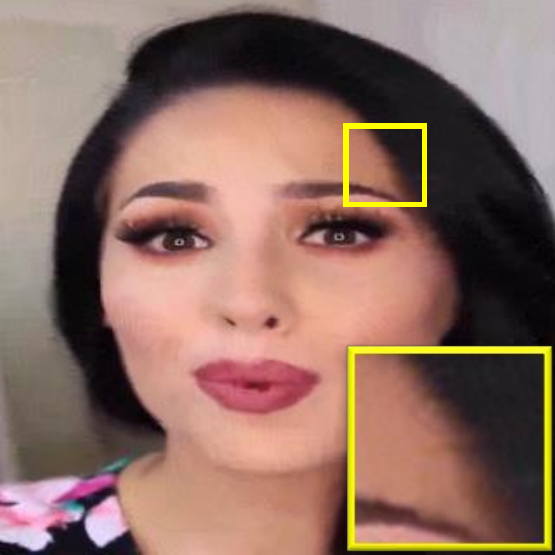}
	\includegraphics[width=\swfive]{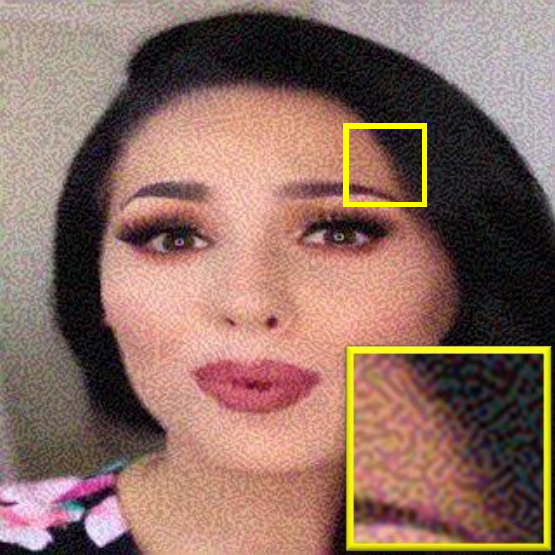}
	\includegraphics[width=\swfive]{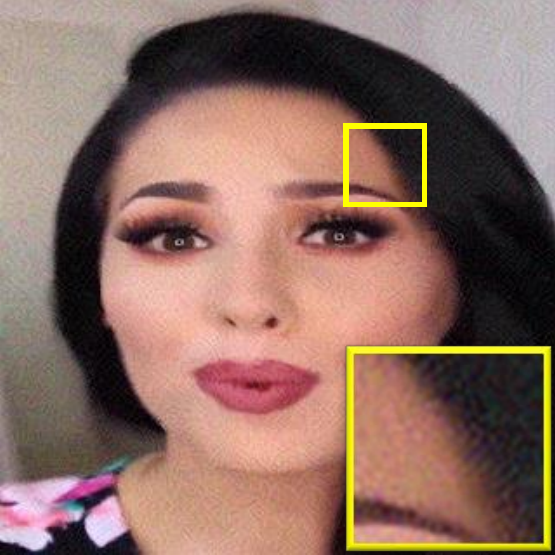}
	\includegraphics[width=\swfive]{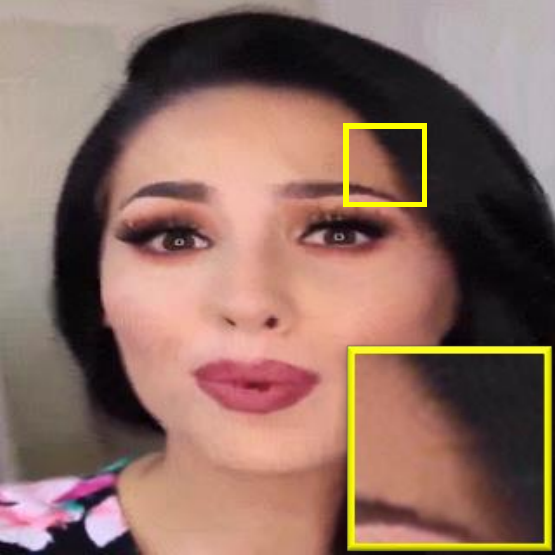}\\
	(a) Original \ \ \ \ (b) FGSM~\cite{goodfellow-2014-fgsm} \ \ \ \  (c) PGD~\cite{madry-iclr18-pgd} \ \ \ \ \ \   (d) Ours\ \ \ \ \  \\
	\vspace{-0.2in}
	\end{center}
\caption{Qualitative evaluation of adversarial examples generated by FGSM~\cite{goodfellow-2014-fgsm}, PGD~\cite{madry-iclr18-pgd} and our method on the FaceForensics++~\cite{rossler-2018-ff++} dataset. These samples contain four types of face forgery generation, i.e., Deepfake, Face2Face, FaceSwap, and NeuralTextures. Although all adversarial examples fool the detectors as real faces successfully, the ones crafted by our hybrid adversarial attack obtain a superior image quality.}
\label{fig:more}
\vspace{-0.05in}
\end{figure}

\section{Conclusion}
In this paper, we propose a frequency adversarial attack method for face forgery detection, which achieves a better image quality compared to the spatial attacks. To further improve its generalization, we propose a hybrid adversarial attack associated with the attacks both in the spatial domain and the frequency domain. The combination of multiple domains reserves their virtues and achieves favorable attack performance both on spatial-based and frequency-based face forgery detectors. Extensive experiments on two datasets indicate that the proposed method not only attacks the white-box models successfully but also enhances the transferability on other models under black-box settings. We hope that our work can draw more attention to the robustness of face forgery detectors.\\

\vspace{-0.10in}

\noindent \textbf{Acknowledgements.} This work was supported by NSFC (61906119, U19B2035), Shanghai Municipal Science and Technology Major Project (2021SHZDZX0102), and CCF-Tencent Open Research Fund.

\clearpage
{\small
\bibliographystyle{ieee_fullname}
\bibliography{egbib}
}

\end{document}